%% file: main.tex
\documentclass[10pt,twocolumn,letterpaper]{article}

\usepackage{cvpr}              
\usepackage[accsupp]{axessibility}
\input{preamble}

\definecolor{cvprblue}{rgb}{0.21,0.49,0.74}
\usepackage[pagebackref,breaklinks,colorlinks,allcolors=cvprblue]{hyperref}

\def\paperID{8554} 
\def\confName{CVPR}
\def\confYear{2026}

\title{MPM: Mutual Pair Merging for Efficient Vision Transformers}

\author{Simon Ravé${}^1$ \quad Pejman Rasti${}^{1,2}$ \quad David Rousseau${}^{1,2}$\\
${}^1$LARIS University of Angers \quad ${}^2$UMR INRAe-IRHS \\
Angers, France\\
{\tt\small \{simon.rave,pejman.rasti,david.rousseau\}@univ-angers.fr}
}

\begin{document}
\maketitle
\input{sec/0_abstract}    
\input{sec/1_intro_real}
\input{sec/2_related_work}
\input{sec/3_method}
\input{sec/4_results}
\input{sec/5_discussion}

\section{Acknowledgements}
This research was funded by the European Union's Horizon Europe Research and Innovation Programme under PHENET project, Grant Agreement No. 101094587. This work was granted access to the HPC resources of IDRIS under the allocation 2024-AD010115553 made by GENCI.

{
    \small
    \bibliographystyle{ieeenat_fullname}
    \bibliography{main}
}


\end{document}

%% file: preamble.tex
\usepackage{microtype}

\renewcommand{\paragraph}[1]{\vspace{.2em}\noindent\textbf{#1.}}

\usepackage{soul}
\setuldepth{foobar}

\usepackage{algorithm}
\usepackage{algorithmic}
\usepackage{caption}
\usepackage[table]{xcolor}

%% file: sec/0_abstract.tex
\begin{abstract}
Decreasing sequence length is a common way to accelerate transformers, but prior token reduction work often targets classification and reports proxy metrics rather than end-to-end latency. For semantic segmentation, token reduction is further constrained by the need to reconstruct dense, pixel-aligned features, and on modern accelerators the overhead of computing merge maps can erase expected gains. We propose Mutual Pair Merging (MPM), a training-free token aggregation module that forms mutual nearest-neighbor pairs in cosine space, averages each pair, and records a merge map enabling a gather-based reconstruction before the decoder so that existing segmentation heads can be used unchanged. MPM introduces no learned parameters and no continuous compression knob (no keep-rate or threshold). The speed-accuracy trade-off is set by a discrete insertion schedule. We benchmark end-to-end latency on an NVIDIA H100 GPU (with and without FlashAttention-2) and a Raspberry Pi 5 across standard segmentation datasets. On ADE20K, MPM reduces per-image latency by up to 60\% for ViT-Tiny on Raspberry Pi 5, and increases throughput by up to 20\% on H100 with FlashAttention-2 while keeping the mIoU drop below 3\%. These results suggest that simple, reconstruction-aware, training-free token merging can translate into practical wall-clock gains for segmentation when overhead is explicitly accounted for.
\end{abstract}

%% file: sec/1_intro_real.tex
\section{Introduction}
\label{sec:intro}

Vision Transformers (ViTs) achieve strong accuracy for semantic segmentation, but their self-attention cost scales quadratically with the number of image tokens, making inference expensive as resolution increases.
A natural response is to reduce the sequence length inside the encoder.
In practice, however, the end-to-end benefit of token reduction is highly hardware- and kernel-dependent: on modern GPUs with highly optimized attention kernels, the additional work needed to compute and apply merge maps can erase or even reverse the expected savings, while on edge CPUs with limited parallelism, reducing tokens often translates directly into latency gains.
This tension motivates a segmentation-oriented token reduction method that is simple enough to deploy, but evaluated and characterized in terms of real wall-clock behavior, including overhead. \cite{dao2022flashattentionfastmemoryefficientexact,dao2023flashattention2fasterattentionbetter,shah2024flashattention3fastaccurateattention} \, \cite{dosovitskiy2020vit,vaswani2017attentionisallyouneed}

Most prior token reduction work targets classification or settings where only a small subset of tokens (for example a class token) is consumed by the prediction head.
Dense prediction is stricter: segmentation decoders typically require a feature at every original patch location, so a reduction step must preserve a faithful mapping back to the full token grid.
Several segmentation-oriented approaches address this with learned policies, locality constraints, or multi-stage schedules, but their reported efficiency is often presented in FLOPs or on a narrow set of accelerator regimes, while the practical overhead introduced by merging, reconstruction, and batching is not consistently quantified. \cite{strudel2021segmenter,cheng2022maskedattentionmasktransformeruniversal,xie2021segformer,zhang2022segvit,bolya2023tokenmergingvitfaster,lu2023contentawaretokensharingefficient,norouzi2024algmadaptivelocalthenglobaltoken}

We introduce \emph{Mutual Pair Merging (MPM)}, a training-free token merging module designed for plug-and-play deployment in ViT-based segmentation pipelines.
MPM computes cosine affinities between tokens, forms pairs using a deterministic \emph{mutual nearest-neighbor} rule, and merges each accepted pair by simple averaging.
A lightweight integer merge map is stored and composed across multiple insertions, and we reconstruct the original $H/P \times W/P$ token grid via a gather-based copy-back before the decoder, so the segmentation head remains unchanged.
MPM has no learned parameters and no continuous compression knob (no keep-rate or threshold). The only user choice is a discrete insertion schedule (how many insertions, and at which depths), which controls the speed-accuracy trade-off.
Unless stated otherwise, we use 0-based indexing and insert MPM before blocks 2 and 5 (the 3rd and 6th blocks), which is the configuration used in our main experiments. \cite{bolya2023tokenmergingvitfaster} \, \cite{strudel2021segmenter}

A key question is whether an additional similarity-and-pairing pass remains worthwhile when attention is already highly optimized.
To address this directly, we report end-to-end wall-clock measurements that \emph{include} MPM overhead, and we separate merge and reconstruction time from the backbone runtime.
In the standard ViT/16 segmentation regime (e.g., $512^2$ inputs, $N=1024$ tokens), the measured merge+reconstruction overhead is small relative to the savings from processing subsequent blocks at reduced sequence length, yielding a large net speedup.
At higher resolutions and with FlashAttention-2 enabled, the net gain is smaller but remains non-negative in our measurements.
We further analyze the content-adaptive nature of MPM by reporting token-count statistics and batch-level effects, since variable sequence lengths can require padding and impact throughput in realistic batched inference. \cite{dao2023flashattention2fasterattentionbetter}

Finally, while MPM performs global pairing in feature space, dense prediction raises a legitimate concern: merging distant but similar regions could blur boundaries or harm small objects.
We therefore characterize the spatial locality of the merges induced by MPM and observe that, in practice, most accepted pairs occur between nearby patches, despite the absence of an explicit locality constraint.
We treat boundary smoothing as an inherent limitation of token aggregation methods in segmentation and quantify where accuracy is lost when compression becomes aggressive.

We evaluate ViT-T/16 through ViT-L/16 Segmenter models on ADE20K, Cityscapes, and Pascal Context, reporting mIoU and end-to-end throughput across an edge CPU (Raspberry Pi 5) and modern GPUs (including H100 with and without FlashAttention-2).
Across these regimes, MPM provides consistent Pareto trade-offs: early insertions yield larger speedups with larger accuracy loss, while late insertions can be close to accuracy-neutral with smaller gains.

\paragraph{Contributions}
\begin{itemize}
\item \textbf{MPM for dense prediction:} a training-free token merging module with deterministic mutual-NN pairing and a reconstruction mapping that preserves decoder compatibility without modifying the segmentation head.
\item \textbf{Decision-grade efficiency evidence:} end-to-end wall-clock evaluation \emph{including} merge and reconstruction overhead, and analysis of when token reduction remains beneficial under optimized attention kernels.
\item \textbf{Trade-off characterization:} depth sweeps and batching-related statistics that make the speed-accuracy behavior reproducible and interpretable for deployment.
\end{itemize}

%% file: sec/2_related_work.tex
\section{Related Work}
\label{sec:related-work}

Reducing inference cost in ViTs typically follows one (or a combination) of three directions: (i) token reduction (selection, pruning, routing, or aggregation/merging) \cite{bolya2023tokenmergingvitfaster,norouzi2024algmadaptivelocalthenglobaltoken,lu2023contentawaretokensharingefficient,rao2021dynamicvit,xu2021evovit,liang2022evit}, (ii) hierarchical encoders that downsample features \cite{liu2021swintransformerhierarchicalvision,ryali2023hierahierarchicalvisiontransformer,wang2021pyramid,wang2022pvt,fan2021multiscale}, and (iii) kernel/compiler-level acceleration such as FlashAttention \cite{dao2022flashattentionfastmemoryefficientexact,dao2023flashattention2fasterattentionbetter,shah2024flashattention3fastaccurateattention}. In semantic segmentation, these choices interact with decoders \cite{strudel2021segmenter,cheng2022maskedattentionmasktransformeruniversal,xiao2018unifiedperceptualparsingscene}, where dense outputs require careful boundary preservation and, if the sequence length is reduced, an “unmerge” mapping back to the original image token grid. Our work belongs to the token reduction family and targets training-free deployment in plain ViTs for segmentation.

\subsection{Token selection: pruning, sampling, routing.}

Dynamic token selection methods estimate token importance and reduce the sequence on the fly. DynamicViT prunes tokens using lightweight predictors learned end-to-end \cite{rao2021dynamicvit}. EViT preserves attentive tokens and fuses less informative ones guided by class attention \cite{liang2022patchesneedexpeditingvision}. ATS introduces a differentiable, parameter-free sampler that scores tokens using class attention and samples them via inverse transform sampling \cite{fayyaz2022ats}. Other works frame token reduction as budgeted halting (A-ViT) \cite{yin2022avit}, latency-aware soft pruning (SPViT) \cite{spvit2022}, sample-adaptive thresholds (AS-ViT) \cite{liu2023adaptivesparsevitlearnable}, or slow-fast token evolution (Evo-ViT) \cite{xu2021evovit}. Token Cropr \cite{bergner2025token} trains auxiliary cross-attention pruning heads to select task-relevant tokens and removes them at inference, achieving up to 4x speedups with minimal accuracy loss across tasks including ADE20K segmentation.
These approaches typically require fine-tuning and, because keep rates vary per input, batched inference often needs padding or masking, which complicates kernel fusion and can diminish wall-clock gains even when FLOPs drop.

\subsection{Token aggregation: pooling, merging, fusion.}

Aggregation methods reduce tokens by combining them. ToMe shows that pairing tokens via (soft) bipartite matching and averaging can repeatedly halve sequence length in standard ViTs with small accuracy loss and no retraining, yielding large throughput gains on images and videos \cite{bolya2023tokenmergingvitfaster}. Related efforts include Token Pooling \cite{marin2021tokenpooling} and TokenLearner \cite{ryoo2022tokenlearner}, which learn compact token sets, as well as clustering/merging strategies such as Agglomerative Token Clustering (ATC) \cite{haurum2024atc}.
Hybrid schemes bridge pruning and merging, such as ToFu \cite{kim2024tofu} and PPT \cite{wu2024ppt}, arguing that the operations are complementary. Aggregation's strength is portability—training-free, drop-in modules can be added to off-the-shelf backbones. Its main challenge lies in ensuring low overhead and faithful reconstruction for the dense output head.

Several recent works learn data-adaptive merging policies to improve fidelity. DTEM \cite{lee2024learningmergetokensdecoupled} learns a decoupled embedding for merging via a differentiable relaxation, separate from the ViT feature stream. DTMFormer \cite{10.1609/aaai.v38i6.28394} introduces a plug-and-play dynamic token-merging block aimed at segmentation (evaluated in medical imaging), combining attention-guided clustering with reconstruction. These methods typically require fine-tuning and can introduce additional layers or trainable parameters, trading simplicity for improved structure preservation.

\subsection{Token reduction for semantic segmentation.}

Dense prediction adds constraints beyond classification: token reduction must preserve thin structures and fine boundaries, preserve enough tokens for a good reconstruction, and support precise mapping back to pixels. Two segmentation-specific lines of work are most relevant to us:

(i) Segmentation-oriented merging schedules. ALGM proposes a local-then-global schedule on plain ViTs: early local window merges consolidate redundant neighbors, followed mid-network by global bipartite matching. The schedule, justified by depth-wise similarity analysis, can be adjusted at inference for quality-efficiency trade-offs \cite{norouzi2024algmadaptivelocalthenglobaltoken}.
The global bipartite matching follows \cite{bolya2023tokenmergingvitfaster}, constructing a bipartite graph between two equally sized sets of tokens and merging the second set into the first based on the most similar edges. Other methods, such as ATC, report gains for dense tasks through hierarchical agglomerative merging \cite{haurum2024atc}.
ALGM is designed for training-time application to obtain the best performance and compute optimal merging thresholds; however, it still reports competitive results when applied directly without retraining.
Our method shares ALGM's focus on segmentation but with a different design: completely training-free, no locality constraints, and mutual-nearest-neighbor (MNN) pairing averaged symmetrically, with an unmerge map to reconstruct pixel-aligned features for Segmenter.

(ii) Content-aware token sharing. CTS trains a policy network to decide when groups of patches share a token, achieving substantial token reduction without hurting mIoU but at the cost of an extra predictor and two-stage training. The approach is tailored to segmentation and is less drop-in \cite{lu2023contentawaretokensharingefficient}.
Our goal differs: a drop-in, parameter-free merging method that is architecture-agnostic with sufficiently low overhead to remain portable across backbones and hardware.

\subsection{Hardware and compiler awareness.}

Operator-level advances like FlashAttention (v1/2/3) substantially boost attention efficiency on modern GPUs, especially since Hopper, by reducing memory traffic, increasing parallelism, and leveraging low-precision asynchronous pipelines \cite{dao2022flashattentionfastmemoryefficientexact,dao2023flashattention2fasterattentionbetter,shah2024flashattention3fastaccurateattention}. As a result, the value of token reduction becomes backend-dependent: dynamic sequence shortening can disrupt kernel specialization and batched throughput when keep rates vary across samples, causing padding and potential graph recompilation. Similar concerns appear in ToMe's discussion of dynamic lengths and batching \cite{bolya2023tokenmergingvitfaster}.

Our MPM targets low overhead and stable accuracy-speed trade-offs: it computes merge similarity directly on token sequences (allowing insertion before any transformer layer) and keeps an explicit unmerge map for decoding. We also report end-to-end latency—including merge cost—on Raspberry Pi 5 (batch 1/2) and H100 (batch 32), with and without FlashAttention-2 \cite{dao2023flashattention2fasterattentionbetter}, to reflect practical deployment.

\paragraph{Positioning} Training-free merging is attractive for portability. However, wall-clock gains depend on both kernel efficiency and merge overhead. ToMe suggests performing merging between attention and MLP blocks for accuracy, using features within attention \cite{bolya2023tokenmergingvitfaster}. In contrast, we intentionally compute similarity directly on tokens to keep our overhead small and versatility high, accepting a slightly more conservative compression-accuracy curve.
We also acknowledge a common failure mode across token-reduction methods: extreme compression harms thin structures, especially in semantic segmentation, where boundaries are the main vector for prediction performance. We therefore emphasize adaptive keep rates, showcase results across multiple insertion points, and report mIoU / FPS / GFLOPs on ADE20K, Cityscapes, and Pascal Context.

%% file: sec/3_method.tex
\section{Method}
\label{sec:method}

\begin{figure*}[t]
  \centering
  \includegraphics[width=\textwidth]{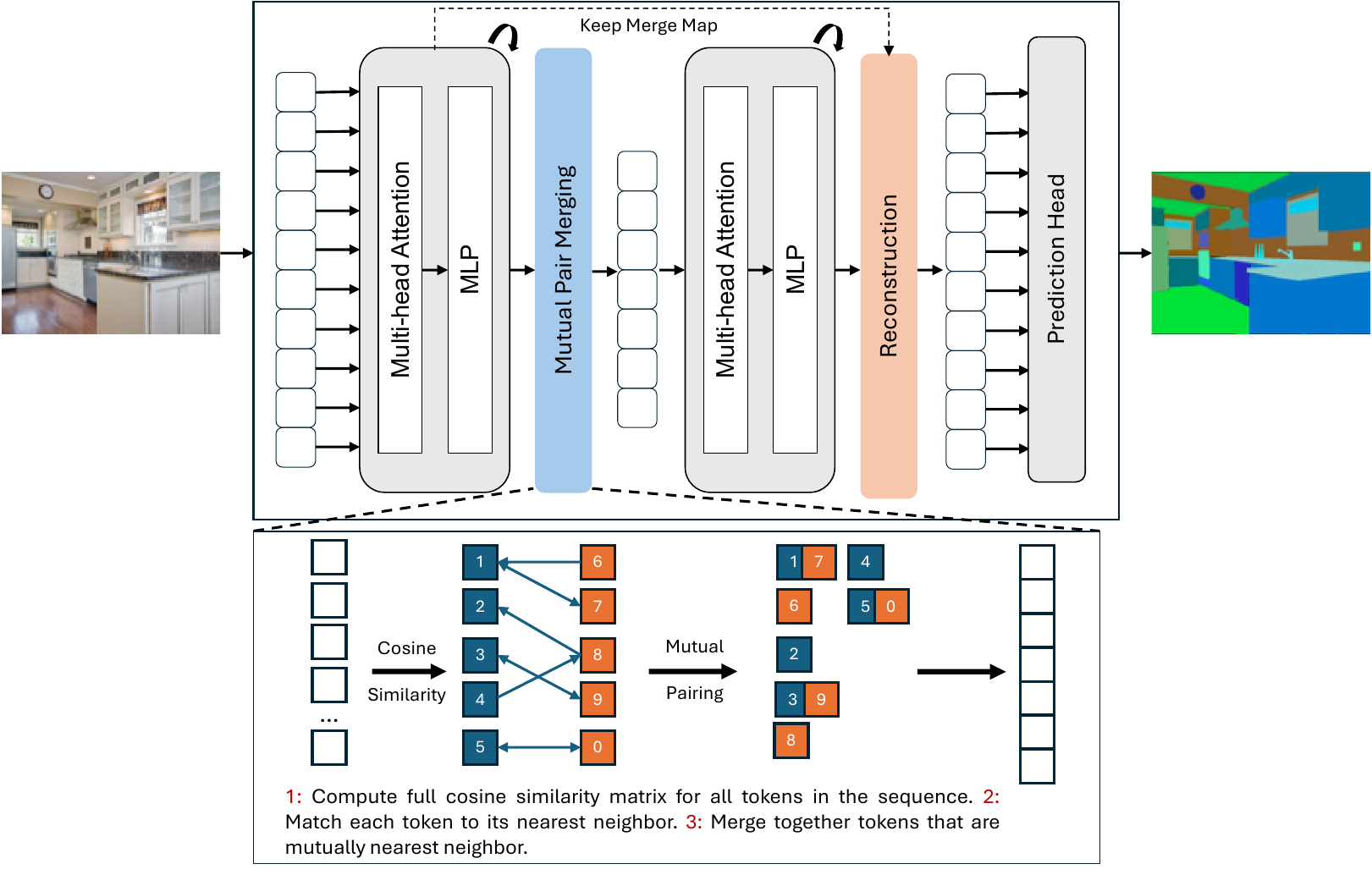}
  \caption{Visual abstract of Mutual Pair Merging (MPM). Similar tokens are matched by mutual pairs and averaged together. Tokens without a mutual relationship remain as singletons. A merge map is kept in memory to enable reconstruction of the entire token sequence. MPM has no learned parameters and no continuous compression knob; the speed-accuracy trade-off is controlled by the choice of insertion blocks.}
  \label{fig:method}
\end{figure*}

We propose \emph{Mutual Pair Merging}, a plug-and-play, training-free module that reduces the number of image tokens in ViT \cite{dosovitskiy2020vit} encoders by merging the most similar \emph{mutual} pairs according to cosine affinity. In our default configuration, MPM is inserted before encoder blocks 2 and 5 (0-based indexing) and leaves special tokens (e.g., class tokens used by the decoder) untouched. The merged tokens propagate through the backbone, and a saved merge map enables exact restoration of the original sequence length via a single gather operation before the decoder, making the method reconstruction-aware and head-agnostic.

\subsection{Preliminaries: ViT Encoder and Mask Transformer Decoder}
\label{sec:prelim}

\paragraph{Backbone}
We adopt a vanilla ViT \cite{dosovitskiy2020vit} with a patch size of $16$ and model scales ViT-T/S/B/L. An input image of spatial size $H\times W$ is partitioned into $P\times P$ patches ($P\!=\!16$), linearly projected to $d$-dimensional tokens, and added to learned absolute positional embeddings. Let $N=\frac{H}{P}\frac{W}{P}$ denote the number of \emph{image} tokens (we use the term ``image tokens'' to distinguish them from any extra/special tokens). Denote the token matrix at the encoder input by $X_0\in\mathbb{R}^{N\times d}$. A transformer block consists of multi-head self-attention (MSA) and a feed-forward network (FFN) with residual connections and layer normalization. We write $X_{\ell+1}=\mathrm{Block}_\ell(X_\ell)$ for block index $\ell$.

\paragraph{Decoder}
For dense prediction, we use the \emph{Mask Transformer} decoder variant of Segmenter \cite{strudel2021segmenter}. The decoder expects the full set of per-patch features arranged on the original $H/P \times W/P$ grid. We therefore apply token merging only inside the encoder and explicitly re-expand the token sequence to length $N$ before passing it to the decoder. No architectural changes to the decoder are required.

We write $X\in\mathbb{R}^{N\times d}$ for a set of image tokens, $d$ is the hidden dimension, and $N$ the number of image tokens at the current point in the network. We let $E$ denote the number of extra/special tokens (e.g., class tokens) that are kept separate and never merged. Batches are processed independently, for clarity we first present the per-image case, ignoring the batch dimension.

\subsection{Mutual Pair Merging}
\label{sec:mpm}

Given image tokens $X=[x_1;\ldots;x_N]\in\mathbb{R}^{N\times d}$, MPM performs three steps: (i) \textbf{cosine similarity}, (ii) \textbf{mutual nearest-neighbor pairing}, and (iii) \textbf{representative averaging}. The procedure is parameter-free and deterministic.

\paragraph{Cosine similarity}
We L2-normalize tokens along the feature dimension and compute the dense cosine affinity:
\begin{equation}
\tilde{X}=\mathrm{norm}_2(X),\qquad
S=\tilde{X}\tilde{X}^\top \in \mathbb{R}^{N\times N},
\end{equation}
with $S_{ii}$ masked to $-\infty$ to prevent self-matching.

\paragraph{Mutual nearest-neighbor pairs}
For each token $i$, let $b(i)=\arg\max_{j\neq i} S_{ij}$ be its most similar neighbor. A \emph{mutual} pair occurs when $b(b(i))=i$ and $b(i)\neq i$. Let $\mathcal{P}=\{(i,j): j=b(i),\, i=b(j),\, i<j\}$ denote the set of undirected mutual pairs, selecting the lower index as the representative of each pair. Tokens not appearing in any pair are treated as singletons.

\paragraph{Representatives and merge operator}
The token set is partitioned into clusters $\{\mathcal{C}_k\}_{k=1}^{N'}$, where each cluster is either a singleton $\{i\}$ or a mutual pair $\{i,j\}\in\mathcal{P}$. Let $r:\{1,\ldots,N\}\to\{1,\ldots,N'\}$ map every token index to its cluster's \emph{compact} representative ID (constructed by a left-to-right scan over the sequence; details below). The merged token matrix $X'\in\mathbb{R}^{N'\times d}$ is the average of members in each cluster:
\begin{equation}
x'_k=\frac{1}{|\mathcal{C}_k|}\sum_{i\in \mathcal{C}_k} x_i, \qquad k=1,\ldots,N',
\end{equation}
with $|\mathcal{C}_k|\in\{1,2\}$. By construction $N' \le \lfloor N/2 \rfloor + (N \bmod 2)$; in practice, not all tokens find mutual partners, so the realized reduction is data-dependent.

\paragraph{Determinism and conflict handling}
Because we only accept mutual pairs and break directionality by retaining the lower index as the representative, ties and conflicts are resolved deterministically without graph heuristics. Tokens chosen by multiple neighbors but lacking reciprocity remain singletons.

\subsection{Insertion into the Backbone}
\label{sec:insertion}

We insert MPM \emph{before} the third and the sixth encoder blocks (index 2 and 5), operating on the block input tokens. Let $X_0$ be the initial image-token matrix (special tokens are excluded from $X_0$). The first MPM at $\mathrm{Block}_2$ produces $(X_1, r^{(1)})$ with $N_1\!=\!|X_1|\le N$. At $\mathrm{Block}_5$, the second MPM operates on the resulting image tokens to produce $(X_2, r^{(2)})$ with $N_2\le N_1$. Subsequent blocks consume $X_2$ until the end of the encoder. Special tokens (count $E$) are concatenated in front of the image tokens throughout and are never considered by MPM.

\paragraph{Composing multiple merge maps}
Each MPM call returns an integer vector of cluster IDs (our $r$). For two stages, the map from original tokens to final representatives is the composition
\begin{equation}
r^{(*)}(i) \;=\; r^{(2)}\!\big(r^{(1)}(i)\big), \qquad i\in\{1,\ldots,N\}.
\label{eq:composition}
\end{equation}
In practice, composition is implemented with a single indexing operation on integer tensors. We store $r^{(*)}$ for reconstruction.

\subsection{Reconstruction for Dense Prediction}
\label{sec:reconstruction}

Let $Z\in\mathbb{R}^{(E+N')\times d}$ be the final encoder output after all blocks, where the first $E$ rows are special tokens (unchanged by MPM) and the last $N'$ rows correspond to merged image tokens. We restore the original image-token sequence length $N$ by \emph{gathering} from the merged set using the composed map $r^{(*)}$:
\begin{equation}
\underbrace{Z_{\mathrm{img}}^{\uparrow}}_{\in \mathbb{R}^{N\times d}}[i] 
\;=\; Z_{\mathrm{img}}\big[r^{(*)}(i)\big], 
\quad i=1,\ldots,N,
\end{equation}
where $Z_{\mathrm{img}} = Z[E{:}E{+}N',: ]$ are the image rows. The final sequence passed to the Mask Transformer decoder is
\[
\big[\,Z_{\mathrm{spec}};\; Z_{\mathrm{img}}^{\uparrow}\,\big]\in\mathbb{R}^{(E+N)\times d},
\]
which exactly matches the decoder's expected input length and preserves the original raster order, enabling a straightforward reshape back to the $(H/P)\times(W/P)$ grid.
Reconstruction is a pure copy; this ``copy-back'' design ensures that the decoder sees the same input shape as in the full-token model.






\subsection{How to choose insertion blocks?}
Early blocks induce the largest token trajectory changes, so reducing tokens there yields the largest latency benefit. Later blocks exhibit smaller token drift; inserting MPM late typically gives smaller speedups with very small accuracy changes. We therefore default to one early and one mid insertion, which empirically provides a favorable accuracy-latency trade-off across CPUs and GPUs. In \cref{fig:ablate-insertion}, we study the impact of different insertion points and choose to insert MPM before blocks 2 and 5. Using two MPM modules yields stronger compression and therefore larger latency gains. MPM has no continuous compression knob, so the speed-accuracy trade-off is controlled by the discrete choice of how many insertions to use and where to place them.

\subsection{Hardware acceleration compatibility}

Because MPM is inserted between transformer blocks, it is fully compatible with existing attention kernel optimizations. Additionally, our goal is to demonstrate improvements across a wide range of hardware, from Raspberry Pi to H100. This is why we implemented our method on ViTs with FlashAttention-2 \cite{dao2023flashattention2fasterattentionbetter}.

\subsection{Why no thresholds or continuous compression knobs?}

\begin{figure}
  \centering
  \includegraphics[width=0.88\columnwidth]{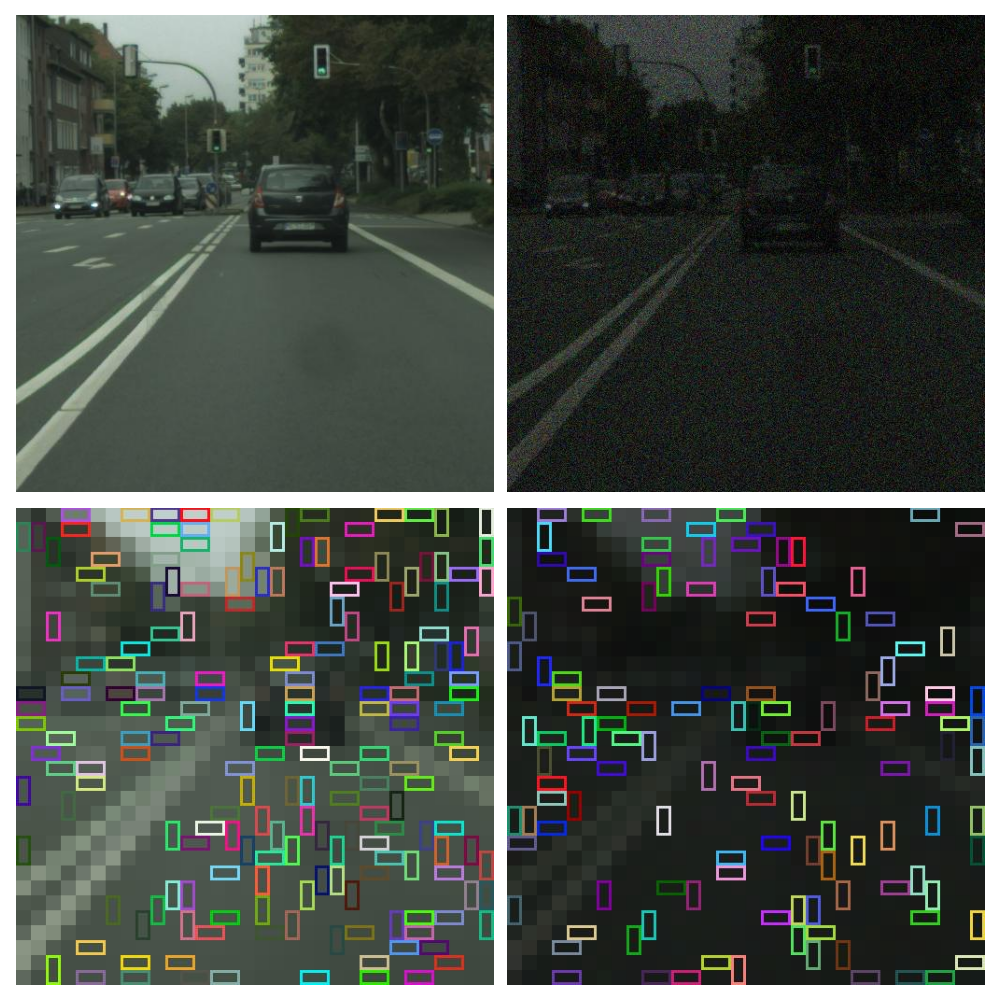}
  \caption{Visualization of MPM on the same image during daytime and nighttime. Nighttime is simulated by reducing the image's luminosity and adding a small amount of thermal and shot noise modeled by Gaussian and Poisson distributions. Approximately 6\% fewer tokens are merged at night than during the day.}
  \label{fig:vis-mpm}
\end{figure}

MPM was designed with online processing in mind. We wanted a method that could be deployed on both high-end hardware and low-cost computers with minimal tuning. In long-running real-world scenarios, such as monitoring a fixed scene with a Raspberry Pi camera, lighting, weather, and scene statistics can change over time. In such settings, a manually chosen merge threshold or keep rate may not remain appropriate. For this reason, MPM uses no learned parameters and no continuous compression knob. Instead, its behavior is controlled only by the discrete insertion schedule. Some methods such as CTS \cite{lu2023contentawaretokensharingefficient} can be seen as inference-time knob-free because the merging policy is fixed during training, but that policy may still be suboptimal under distribution shift. Other methods such as ToMe \cite{bolya2023tokenmergingvitfaster} use a fixed merge ratio, which has advantages for batching, but still requires manual adjustment when scene conditions change. We visualize this behavior in \cref{fig:vis-mpm}.

\subsection{Algorithm}

\label{sec:pseudocode}

We provide high-level pseudocode for a single MPM call in \cref{alg:mpm}. The implementation is fully vectorized and corresponds directly to the description above. 

\begin{algorithm}[t]
\small
\caption{Mutual Pair Merging}
\label{alg:mpm}
\begin{algorithmic}[1]
\REQUIRE Image tokens $X\!\in\!\mathbb{R}^{N\times d}$
\STATE $\tilde{X} \leftarrow \mathrm{row\_normalize}(X)$
\STATE $S \leftarrow \tilde{X}\tilde{X}^\top$ 
\STATE $b(i) \leftarrow \arg\max_{j\neq i} S_{ij}$ for all $i$
\STATE $\mathcal{P} \leftarrow \{(i,j): b(i)=j, b(j)=i, i<j\}$ 
\STATE $\textsc{RepMask}[i] \leftarrow \mathbf{1}[\,\exists j:(i,j)\in\mathcal{P}\,] \lor \mathbf{1}[\,i \text{ is singleton}\,]$
\STATE Assign compact IDs $r(i)\in\{1,\ldots,N'\}$ by a left-to-right scan over \textsc{RepMask} (representatives get new IDs; partners inherit their representative's ID)
\STATE $x'_k \leftarrow \frac{1}{|\{i:r(i)=k\}|}\sum_{i:r(i)=k} x_i$ for $k=1\ldots N'$
\RETURN $X'=[x'_1;\ldots;x'_{N'}]$, merge map $r$
\end{algorithmic}
\end{algorithm}

\paragraph{Summary}
MPM performs dense, training-free, and deterministic token merging based on mutual nearest neighbors in cosine space. With the default schedule, it is inserted before encoder blocks 2 and 5 (0-based indexing), yields content-adaptive reductions, and preserves compatibility with standard segmentation decoders via a lightweight gather-based reconstruction step. The method has no learned parameters, no continuous compression knob, and is implemented in a single PyTorch function.

%% file: sec/4_results.tex
\section{Results}
\label{sec:Results}

\paragraph{Evaluation protocol}
To evaluate the performance of our method, we implement it on ViT-based segmentation models and compare it to state-of-the-art token reduction methods. When possible, base models are taken from \cite{strudel2021segmenter}, the others are trained using mixed precision on an H100 GPU. All training details are available in the supplementary material.

We first report the main results in \cref{tab:segmenter_comparison} for the ADE20K dataset, following the \texttt{mmseg} evaluation routine.
We compute the standard single-scale mIoU on the validation split. 
All methods use the same Mask Transformer \cite{strudel2021segmenter} decoder and off-the-shelf ViT checkpoints.

\paragraph{Baselines and fairness}
We compare to multiple token-reduction baselines, including ToMe, ALGM, CTS, and the full-token model without merging. 
MPM is inserted before Blocks 2 and 5. 
For all methods, we follow the best hyperparameters reported in their respective papers. 
For latency metrics, the model and method overhead are included in the timings. Dynamic GFLOPs are reported as the mean over one image across the evaluation dataset using the PyTorch profiler.

\paragraph{Hardware and measurement}
GPU throughput is measured on a single \textbf{H100 SXM} with a batch size of \(B{=}32\). We perform 50 warmup steps and time the full validation run with explicit \texttt{torch.cuda.synchronize()}. Edge CPU measurements use a \textbf{Raspberry Pi~5} with 20 warmup steps.

\begin{table}[htbp]
\centering
\begin{tabular}{lccc}
    \toprule
    \textbf{Model} & \textbf{mIoU} & \textbf{GFLOPs} & \textbf{Latency (FPS)} \\
    \midrule
    \textbf{Seg-T/16} & 38.1 & 25 & 660 \\
    ALGM* & 38.9 & $\sim$16.7 & 665 \\
    CTS* & 38.2 & - & 547 \\
    ToMe & 38.1 & $\sim$19 & 751 \\
    \rowcolor{pink!30} MPM (2,5) & 37.6 & $\sim$17.6 &  \textbf{831} \\
    \midrule
    \textbf{Seg-S/16} & 45.3 & 76.9 & 333 \\
    ALGM* & 46.4 & $\sim$55 & 343 \\
    CTS* & 45 & - & 333 \\
    ToMe & 45.5 & $\sim$59 & 388 \\
    \rowcolor{pink!30} MPM (2,5) & 45.1 & $\sim$54 & \textbf{431} \\
    \midrule
    \textbf{Seg-B/16} & 48.5 & 258 & 133 \\
    ALGM* & 49.4 & $\sim$160 & 138 \\
    CTS* & 48.7 & - & 160 \\
    ToMe & 49.1 & $\sim$201 & 160 \\
    \rowcolor{pink!30} MPM (2,5) & 48.0 & $\sim$184 & \textbf{177} \\
    \midrule
    \textbf{Seg-B/8} & 49.5 & 1320 & 20 \\
    \rowcolor{pink!30} MPM (2,5) & 49.2 & $\sim$809 & \textbf{35} \\
    \midrule
    \textbf{Seg-L/16} & 51.7 & 800 & 47 \\
    ALGM* & 52.7 & $\sim$624 & 50 \\
    CTS* & 50.75 & - & 66 \\
    ToMe & 51.8 & $\sim$633 & 56 \\
    \rowcolor{pink!30} MPM (2,5) & 50.4 & $\sim$496 & \textbf{74} \\
    \bottomrule
\end{tabular}
\caption{Comparison of main token reduction methods on ADE20K; images are 512x512 pixels. Results were recorded on a single H100 GPU in full precision without FlashAttention. * indicates methods that require training or fine-tuning, and $\sim$ denotes the average recorded FLOPs per image over the entire dataset with random batches.}
\label{tab:segmenter_comparison}
\end{table}

\begin{table}[ht]
\centering
\begin{tabular}{lccc}
\hline
\textbf{Model} & \textbf{mIoU} & \textbf{GFLOPs} & \textbf{FPS (B=32)} \\
\toprule
\textbf{ViT-S/16} & 53.0 & 64 & 426 \\
CTS* & 52.7 & - & 438 \\
ALGM* & 53.2 & $\sim$52 & 430 \\
ToMe  & 53  & $\sim$45    & 505  \\
\rowcolor{pink!30} MPM (2,5) & 52.5  & $\sim$44 & \textbf{561} \\
\midrule
\textbf{ViT-B/16} & 55.0 & 219 & 159 \\
CTS* & 54.5 & - & 197 \\ 
ALGM* & 55 & $\sim$164 & 166 \\
ToMe & 54.8 & $\sim$161 &  201 \\
\rowcolor{pink!30} MPM (2,5) & 54.5 & $\sim$153 &  \textbf{213} \\
\midrule
\textbf{ViT-L/16} & 58.0 & 686 & 56 \\
CTS* & 57.09 & - & 79 \\ 
ALGM* & 58.0 & $\sim$400 & 58 \\
ToMe & 58 & $\sim$522 &  69 \\
\rowcolor{pink!30} MPM (2,5) & 57.3 & $\sim$416 & \textbf{84} \\
\bottomrule
\end{tabular}
\caption{Results on Pascal Context dataset (H100, Float 32) with Mask Transformer segmentation head \cite{strudel2021segmenter}, 480x480px per image, base sequence length of 900 tokens for patch size 16.}
\label{tab:pcontext_results}
\end{table}

\begin{table}[ht]
\centering
\begin{tabular}{lccc}
\hline
\textbf{Model} & \textbf{mIoU} & \textbf{GFLOPs} & \textbf{FPS (B=32)} \\
\toprule
\textbf{ViT-S/16} & 76.3 & 116 & 106 \\
 CTS* & 76.6 & - & \textbf{169} \\
 ALGM* & 76.9 & 79 & 132 \\
 ToMe & 76.3 & $\sim$97 & 105 \\
\rowcolor{pink!30} MPM (2,5) & 75.3 & $\sim81$ & 158 \\
\midrule
\textbf{ViT-B/16} & 77.3 & 348 & 44 \\
 CTS* & 78.11 & - & \textbf{80} \\
 ALGM* & 76.4 & $\sim211$ & 62 \\
 ToMe & 77.3 & $\sim$298 & 46 \\
\rowcolor{pink!30} MPM (2,5) & 76.3 & $\sim247$ & 65 \\
\bottomrule
\end{tabular}
\caption{Results on the Cityscapes dataset (H100, float32) with the Mask Transformer segmentation head \cite{strudel2021segmenter}. * indicates methods that require training or fine-tuning. Input images were 768x768, corresponding to 2304 tokens in the base sequence.}
\label{tab:cityscapes_results}
\end{table}

\begin{table}[h!]
\centering
\begin{tabular}{lcc}
\toprule
\textbf{Model} & \textbf{FPS B=1} & \textbf{FPS B=2} \\ 
\midrule
\textbf{ViT-T/16}  & 1.06 & 1.05 \\
ALGM*              & 1.64 & 1.47 \\
\rowcolor{pink!30} MPM (2,5) & \textbf{1.71} & \textbf{1.75} \\
\midrule
\textbf{ViT-S/16} & 0.49 & 0.44 \\
ALGM*              & 0.69 & 0.64 \\
\rowcolor{pink!30} MPM (2,5) & \textbf{0.73} & \textbf{0.71} \\
\bottomrule
\end{tabular}
\caption{Comparison of latency for merging methods on a Raspberry Pi 5 on the ADE20K dataset.}
\label{tab:fps-raspberry}
\end{table}

\begin{table}[h!]
\centering
\begin{tabular}{l c}
\hline
\textbf{Model} & \textbf{Latency Batch Size = 32 (FPS)} \\
\toprule
\textbf{ViT-B/16} & 735 \\
ALGM* & 643 \\
ToMe & 753 \\
\rowcolor{pink!30} MPM (2,5) & \textbf{780} \\
\midrule
\textbf{ViT-L/16} & 375 \\
ALGM* & 351 \\
ToMe & 350 \\
\rowcolor{pink!30} MPM (2,5) & \textbf{456} \\
\bottomrule
\end{tabular}
\caption{Latency comparison (FPS) for different merging methods at batch size 32 on H100 GPU in half precision (BFloat16) using FlashAttention-2 \cite{dao2023flashattention2fasterattentionbetter}.}
\label{tab:latency_fps_flashattn}
\end{table}

\begin{table}[ht]
\centering
\begin{tabular}{lccc}
\toprule
\textbf{Model} & \textbf{mIoU} & \textbf{GFLOPs} & \textbf{FPS (B=32)} \\
\midrule
\textbf{DeiT-B Mask} & 46.2 & 259 & 130 \\
\rowcolor{pink!30} MPM (2,5) & 45.6 & $\sim187$ & \textbf{174} \\
\midrule
\textbf{ViT-B/16 Linear} & 47.8 & 214 & 159 \\
\rowcolor{pink!30} MPM (2,5) & 46.7 & $\sim141$ & \textbf{228} \\
\bottomrule
\end{tabular}
\caption{Results using a DeiT backbone \cite{touvron2021trainingdataefficientimagetransformers} with a Mask Transformer head and a base ViT with a linear head. We observe trends similar to those in \cref{tab:segmenter_comparison}.}
\label{tab:deit_linear_results}
\end{table}


\begin{table}[htbp]
\centering
\begin{tabular}{lccc}
    \toprule
    \textbf{Model} & \textbf{mIoU} & \textbf{FPS (B=4)} \\
    \midrule
    \textbf{EVA-01} & 61.5 & 2.61 \\
    \rowcolor{pink!30} MPM (2,5) & 61.4 & \textbf{3.70} \\
    \bottomrule
\end{tabular}
\caption{Results on ADE20K using EVA-01 \cite{fang2022evaexploringlimitsmasked} backbone with ViT Adapter \cite{chen2023visiontransformeradapterdense} and Mask2Former \cite{cheng2022maskedattentionmasktransformeruniversal} head. Due to memory limitations, latency is measured at batch size 4 on a single A100 GPU.}
\label{tab:eva_results}
\end{table}

\subsection{Ablation} \label{ablation}

\begin{figure}[t]
    \centering
    \includegraphics[width=\columnwidth]{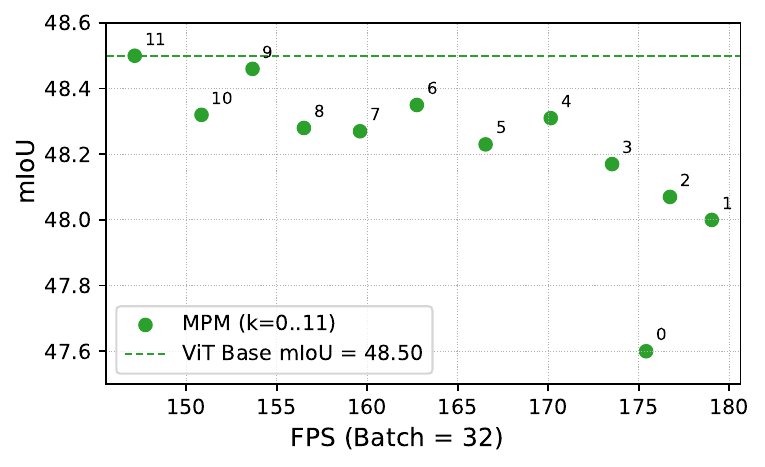}
    \caption{Accuracy vs.\ FPS for MPM using different insertion layers in ViT-Base \cite{dosovitskiy2020vit} on ADE20K \cite{zhou2017ade20k}.}
    \label{fig:ablate-insertion}
\end{figure}

\paragraph{Insertion depth}
In \cref{fig:ablate-insertion}, we compare merging at different depths to quantify the speed-accuracy trade-off.
Early token merging provides strong speedups but incurs modest mIoU drops, while later merging recovers accuracy with lower gains. Merging after layer 5 is nearly ``free'' in accuracy, so we place MPM at layers 2 and 5: layer 2 gives the best speed-accuracy trade-off, and layer 5 adds extra savings with negligible loss. Using two merge points maximizes latency reduction while keeping accuracy loss within \mbox{$\sim$1--2\%}.

\paragraph{Different backbone and head}
In \cref{tab:deit_linear_results}, we apply MPM to segmentation models using DeiT \cite{touvron2021trainingdataefficientimagetransformers} backbones, as well as to plain ViT models with a linear head instead of a Mask Transformer head \cite{strudel2021segmenter}. We find that the results are consistent with those obtained on basic ViT models with a Mask Transformer head. The loss in mIoU is around 2\%, while the gain in latency exceeds 30\%, which matches the main results shown in \cref{tab:segmenter_comparison}.

\vspace{0.25em}

\subsection{Analyses}

On \cref{tab:segmenter_comparison}, MPM provides a clear latency improvement over existing methods on high-end GPUs, giving more than 50\% higher FPS for Seg-L/16 without hardware-specific acceleration. We use random batches of size 32, which introduces realistic variability for adaptive methods. Within a batch, an image that compresses well may still need to be padded to match a harder image whose sequence length is barely reduced, which can negate part of the compression gain. This explains why our ALGM latency results differ from those reported in the original paper. Further details and batch-level standard errors are provided in the supplementary material. Results are consistent across datasets, as shown in \cref{tab:pcontext_results} for Pascal Context and \cref{tab:cityscapes_results} for Cityscapes. However, on Cityscapes, CTS \cite{lu2023contentawaretokensharingefficient} performs better in both accuracy and latency. We suspect this is because CTS is well suited to datasets with larger objects and more clearly separated classes, allowing more aggressive token sharing within spatial clusters.

The accuracy drop of MPM is slightly larger than that of other methods, about 1--2\% mIoU, which is expected since MPM is training-free and does not adapt the backbone to merged tokens. We also use an intentionally aggressive setup with two MPM layers to maximize latency gains. Overall, we consider this a reasonable trade-off given the substantial speedups and MPM's applicability to any pre-trained ViT without retraining or continuous compression tuning. On larger models such as EVA-01 \cite{fang2022evaexploringlimitsmasked}, the performance drop is much smaller, around 0.2\%, despite latency improvements of up to 40\%.

Furthermore, \cref{tab:segmenter_comparison} shows that GFLOPs per image are not directly predictive of latency, especially for dynamic token-reduction methods such as ALGM \cite{norouzi2024algmadaptivelocalthenglobaltoken}. Even under the quadratic complexity of full attention, reducing GFLOPs or token counts by 30\% does not imply an equal latency reduction because not all operations parallelize equally well. This becomes even more apparent with optimized kernels such as FlashAttention \cite{shah2024flashattention3fastaccurateattention}. \cref{tab:latency_fps_flashattn} shows that some existing methods can become slower than the full model once their overhead is included. In contrast, MPM remains faster, although the gains are much smaller than in the non-FlashAttention setting and become modest on small models. With ViT-Tiny, we observe only a 4\% latency improvement, whereas on ViT-Large the FPS increases by 30\%.

On \cref{tab:fps-raspberry} for Raspberry Pi 5, we obtain similar results. We measure a 60\% latency improvement on ViT-Tiny and a 50\% improvement on ViT-Small, which is consistent with the trends observed on high-end GPUs. For ALGM \cite{norouzi2024algmadaptivelocalthenglobaltoken}, we recover the expected latency improvement once the batch size is reduced relative to \cref{tab:segmenter_comparison}. We also observe that increasing the batch size provides no benefit on this edge device because the Raspberry Pi 5 offers limited parallelism.

\subsection{How does it scale?}
The main bottleneck of MPM is the computation of the full similarity matrix between all tokens. For regular sequences (e.g., 1024 tokens), this is not considered a problem, but for higher-resolution images, one might suspect that the similarity computation could increase latency. However, note that MPM achieves good results on the EVA-01 \cite{fang2022evaexploringlimitsmasked} backbone with a Mask2Former head \cite{cheng2022maskedattentionmasktransformeruniversal} in \cref{tab:eva_results}, even though the images are processed at a resolution of 896x896, corresponding to 3136 tokens at a patch size of 16. This demonstrates that MPM can scale to very large models and higher-resolution images while still providing significant latency improvements (over 40\%) with minimal accuracy loss. Additionally, we observe a clear latency improvement when applying MPM to ViT-B/8 in \cref{tab:segmenter_comparison}, even though the model processes 4096 tokens. To further quantify the actual overhead of MPM, we provide detailed per-Transformer block timing comparisons, with and without MPM, in the supplementary materials.

%% file: sec/5_discussion.tex
\section{Conclusion and Discussion}
\label{sec:Discussion}

MPM demonstrates that a training-free, mutual-nearest-neighbor merge can convert token redundancy into real wall-clock gains across heterogeneous hardware while preserving segmentation quality, with small and predictable drops. Unlike methods that rely on thresholds or learned policies, MPM's simplicity and early insertions keep overhead low enough for reductions to materialize as FPS improvements, even on modern GPUs where GFLOPs can be a poor proxy for latency. Our experiments highlight a hardware-dependent trade-off: on edge CPUs, limited parallelism allows reductions to translate directly into speed gains, whereas on highly parallel GPUs the benefit grows with backbone size and depends more strongly on token-reduction overhead.
At the same time, MPM inherits known limitations of token aggregation for dense prediction: small objects and fine boundaries can be slightly smoothed, and the $O(N^2)$ cosine pass, while modest in our standard ViT/16 measurements, may become a bottleneck at very large $N$. To address these limitations, chunking or locality constraints could be applied to the similarity computation.
Additionally, token reduction is only truly effective for full-attention models. For hierarchical backbones \cite{liu2021swintransformerhierarchicalvision,ryali2023hierahierarchicalvisiontransformer}, where attention complexity can become linear with sliding-window mechanisms, reducing the token sequence yields little or no performance gain.
In conclusion, we eliminate the need for retraining and continuous compression tuning, yielding a plug-and-play module for existing ViT+Segmenter setups and related ViT-based architectures. Our results show that, with careful merge-and-expand engineering, token reduction can be an effective inference-time optimization. We hope this motivates further exploration of methods adaptable to diverse hardware.
